\documentclass{article}

\usepackage[utf8]{inputenc} %
\usepackage[T1]{fontenc}    %
\usepackage{graphicx}       %
\usepackage{amsmath}        %
\usepackage{amssymb}        %
\usepackage{amsfonts}       %
\usepackage{url}            %
\usepackage{booktabs}       %
\usepackage{nicefrac}       %
\usepackage{microtype}      %
\usepackage[super,sort&compress]{natbib} %
\bibpunct{}{}{,}{s}{,}{,}  %

\usepackage[margin=1in]{geometry} %
\usepackage{dirtree} %
\usepackage{tcolorbox} %

\usepackage{hyperref}       %
\usepackage{xcolor}         %
\usepackage{cleveref}       %
\usepackage{listings}       %
\hypersetup{
    colorlinks=true,
    allcolors=black,
}

\begin{document}

\title{MegNIST, A Benchmark for Non-Invasive Inner Speech Decoding}

\author{
Teyun Kwon$^{1}$, 
Pratik Somaiya$^{1}$, 
Gereon Elvers$^{1}$, 
Francesco Mantegna$^{1}$, \\
Anna Camera$^{2}$, 
Sven Braeutigam$^{2}$, 
Mark Woolrich$^{2}$, 
Oiwi Parker Jones$^{1,3}$ \\
\\
$^{1}$PNPL\includegraphics[height=1.5\fontcharht\font`\B]{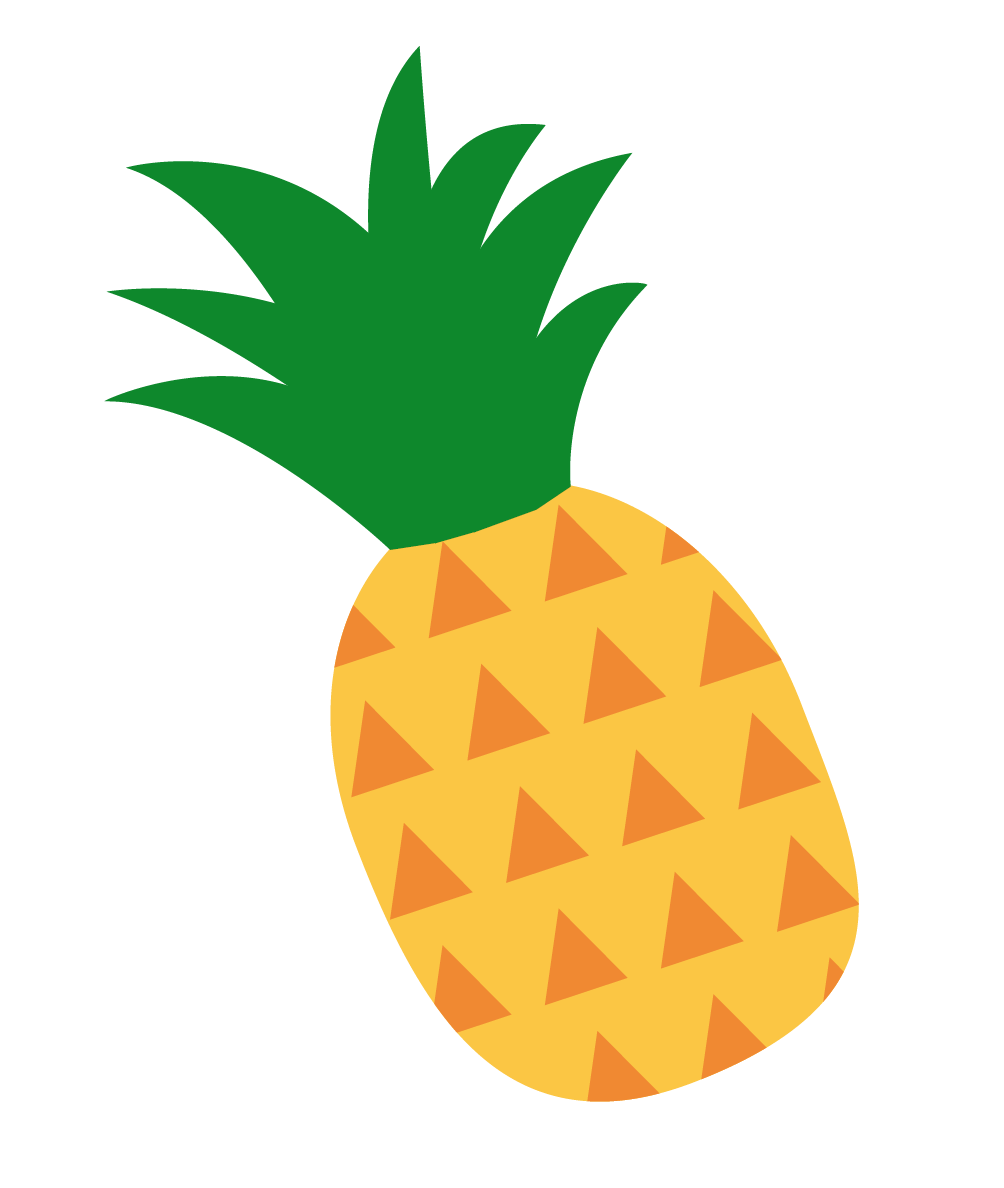}, Department of Engineering Science, University of Oxford, UK \\
$^{2}$OHBA, Wellcome Centre for Integrative Neuroimaging, University of Oxford, UK \\
$^{3}$Jesus College, University of Oxford, UK \\
\\
\texttt{oiwi@robots.ox.ac.uk}
}

\date{} %
\maketitle

\begin{abstract}

MegNIST is a benchmark dataset for developing and evaluating machine learning methods for inner speech decoding from magnetoencephalography (MEG). The dataset comprises 12,000 trials of a single participant imagining speaking the digits zero through nine, recorded across four sessions. Each trial contains 306-channel MEG recordings sampled at 250 Hz, preprocessed using signal space separation, filtering, and downsampling. Data are provided as pre-defined training (10,000 trials), validation (1,000 trials), and test (1,000 trials) splits, with entire experimental runs assigned to each split to preserve temporal independence. The dataset is available in both BIDS format for neuroimaging analysis and as serialised tensors for machine learning workflows. Baseline classification results demonstrate that imagined digits can be decoded significantly above chance. Standardised splits, validated baselines, example code, and a PyTorch-compatible loader in the \texttt{pnpl} Python library enable straightforward integration with machine learning frameworks, lowering barriers for researchers to rapidly develop and benchmark inner speech decoding methods for non-invasive brain-computer interfaces.

\end{abstract}

\section{Background \& Summary}

\subsection{Motivation}

MNIST\cite{lecun1998mnist, lecun2010mnist}, a benchmark dataset of 70,000 handwritten digits (0–9) with predefined train/test splits, has been called the ``machine learning equivalent of fruit flies''\cite{hinton2012coursera}. Much as fruit flies serve biology, MNIST serves as a model system for rapid hypothesis testing and methods development in computer vision. 
The analogy reflects a rich tradition: digits have long served as a testing ground for pattern recognition, from Bell Labs' work in the 1950s\cite{davis1952automatic} to modern deep learning benchmarks. By providing an accessible, standardised dataset, MNIST democratised computer vision, enabling rapid prototyping and onboarding of new researchers. This success catalysed progression from simple benchmarks to larger datasets like CIFAR-10\cite{krizhevsky2009cifar10} and ImageNet\cite{deng2009imagenet}, ultimately enabling the computer vision applications we see today, and inspiring numerous variants including Fashion-MNIST\cite{xiao2017fashion}, Extended MNIST\cite{cohen2017emnist}, Kuzushiji-MNIST\cite{clanuwat2018deep}, and MNIST-1D\cite{greydanus2024mnist-1d}. Similar accessible benchmarks have been established in other domains, such as audio (TIMIT\cite{garofolo1993timit}, LibriSpeech\cite{panayotov2015librispeech}) for speech recognition.

The situation differs markedly for neural decoding---the machine learning task of mapping brain recordings to meaningful outputs. Although neural decoding presents an ideal challenge and emerging subdomain for deep learning, the field lacks accessible benchmark datasets. Several barriers have historically limited accessibility: domain expertise requirements for data access and preprocessing, gated datasets requiring institutional approval, and public datasets hosted on platforms that throttle downloads. LibriBrain\cite{ozdogan2025libribrain} and LibriBrain100\citep{mantegna2026libribrain100} represent notable exceptions, providing large-scale MEG recordings on Hugging Face with automated download through the \texttt{pnpl} library\cite{landau2025competition} (a PyTorch-compatible toolkit for loading and processing brain data). This infrastructure is enabling rapid methods development\cite{elvers2025elementary,jayalath2025unlocking} and powering open machine-learning competitions\cite{landau2025competition, elvers2026benchmarking, mantegna2026pnplcompetition}. However, LibriBrain, LibriBrain100, and other publicly available datasets\cite{schoffelen2019, gwilliams2023megmasc, armeni2022} focus on \textit{heard speech}---participants listening to connected, naturalistic speech such as sentences, podcasts, or audiobooks. Inner speech decoding, by contrast, lacks such foundational benchmarks, limiting progress toward practical non-invasive brain-computer interfaces.

In healthy volunteers, three types of speech can be studied: heard speech (listening to speech), inner or covert speech (silent imagination without vocal articulation), and overt speech (speaking aloud). Invasive brain-computer interfaces using electrocorticography (ECoG) have achieved remarkable progress decoding attempted speech (the neural equivalent of overt speech in individuals unable to physically articulate) in paralysed patients with dysarthria\cite{moses2021neuroprosthesis, willett2023high, card2024nejm}. Non-invasive approaches are vastly preferable, avoiding surgical risks whilst potentially reaching broader populations. In healthy controls, studying overt speech is problematic for BCI development: overt articulation introduces substantial artefacts in non-invasive recordings (movement-related noise in EEG/MEG, correlated motion in fMRI) that complicate interpretation. Moreover, these individuals can already speak. Therefore, inner speech represents a more promising path forward for developing non-invasive BCIs. Here we introduce MegNIST---an MNIST for MEG data---which addresses this need by providing the first standardised benchmark for inner speech decoding, designed to enable rapid prototyping and accelerate methods development in neural decoding research.

\subsection{Related Work}

Among non-invasive neuroimaging modalities, magnetoencephalography (MEG) achieves an optimal balance between temporal and spatial resolution for speech decoding applications\cite{gross2019, dash2020phrases}. MEG combines millisecond temporal precision with superior spatial localisation compared to electroencephalography (EEG), as magnetic fields pass through the skull with less distortion than electrical fields\cite{gross2019}. While functional magnetic resonance imaging (fMRI) provides excellent spatial resolution, its temporal resolution is limited to seconds or tens of seconds\cite{jenkinson+chappell2018neuroimaging}, inadequate for characterising the rapid dynamics of speech processing.

Existing inner speech datasets are limited in scale and standardisation. Early work focused on phoneme classification from EEG\cite{brigham2010imagined, dzmura2009toward, dasalla2009single}. Word-level studies achieved good accuracy (80--91\%) on binary phonological tasks\cite{zhao2015classifying} and classification of limited vocabularies\cite{suppes1997brain, porbadnigk2009eeg}. Foundational MEG work by Tian and Poeppel\cite{tian2010mental, tian2012mental} examined the neural basis of inner speech. However, these studies collected data across multiple participants, introducing inter-subject variability, and lack standardised evaluation protocols (i.e., pre-defined train-validation-test splits with clear guidelines for performance reporting). Whilst valuable for proof-of-concept studies, they are not designed as machine-learning-ready resources for methods development.

The MindBigData collection\cite{vivancos2022mindbigdata, vivancos2023mindbigdata8b} might appear most similar to our dataset, given its explicit MNIST reference and focus on EEG-based digit decoding. However, the original publications provide no baseline models validating that recorded signals contain decodable digit information. A community-maintained leaderboard also reports extreme variance in classification accuracy and significant reproducibility issues, with high-performing methods typically lacking available code and independent validation attempts failing to achieve comparable results (\url{https://huggingface.co/spaces/DavidVivancos/MindBigData-Leaderboard}, accessed 13 June 2025). These challenges underscore the importance of rigorous validation and reproducible baselines. MegNIST provides comprehensive preprocessing documentation, baseline models, and open-source implementations (Sections~\ref{sec:tech_val}--\ref{sec:usage_notes}) to ensure reliability.

A key design decision concerns whether to prioritise breadth (many subjects with limited data each) or depth (extensive data from fewer subjects). Recent findings demonstrate that per-subject data quantity matters more than total dataset size across subjects\cite{dascoli2025natcomm}, motivating a ``deep'' (extensive within-subject) rather than ``broad'' (multi-subject) approach. LibriBrain\cite{ozdogan2025libribrain} and LibriBrain100\cite{mantegna2026libribrain100} exemplify this principle: substantial single-subject data enabled strong performance on downstream tasks including word classification\cite{jayalath2025unlocking}. This deep, within-subject focus addresses the classic neural decoding challenge of accounting for individual variability\cite{csaky2023group}. MegNIST adopts this deep approach with 12,000 trials from a single participant, a scale that balances depth sufficient for deep learning with the accessibility and rapid iteration characteristic of MNIST-style benchmarks.

\subsection{The MegNIST Dataset}

MegNIST provides 12,000 MEG recordings of a single participant imagining the digits zero through nine, evenly distributed across the 10 digit classes. The dataset includes pre-defined training (10,000 trials), validation (1,000 trials), and test (1,000 trials) splits with entire experimental runs assigned to each split rather than randomly shuffling individual trials. This preserves temporal independence and protects against information leakage from slow physiological changes across adjacent trials\cite{harris2020nonsense}, whilst reflecting realistic brain-computer interface deployment where models must generalise to future recording sessions.

Following the examples of LibriBrain and LibriBrain100\cite{ozdogan2025libribrain,mantegna2026libribrain100}, MegNIST prioritises accessibility in both data access and ease of integration with deep learning frameworks. The dataset is hosted on Hugging Face (a widely-used platform for sharing datasets and models) with automated download through the \texttt{pnpl} library\cite{landau2025competition}, enabling straightforward integration with frameworks like PyTorch\cite{paszke2019pytorch}. Data are available in BIDS format\cite{niso2018meg-bids} for neuroimaging analysis and as serialised tensors for machine learning workflows. Baseline classification demonstrates that imagined digits can be decoded from held-out recordings at levels significantly above chance (22.6\% accuracy vs.~10\% chance, $p < 10^{-5}$, permutation test). Complete preprocessing pipelines and baseline models are provided as open-source implementations. %

Like MNIST for computer vision\cite{lecun1998mnist, lecun2010mnist}, MegNIST is designed as an accessible entry point for neural decoding research. The simple digit classification task, standardised splits, validated baselines, and straightforward data access aim to lower barriers to entry and enable rapid iteration on new approaches. By providing a common benchmark, MegNIST facilitates progress measurement and method comparison in inner speech decoding. Initial development in healthy participants establishes foundational methods that can subsequently be adapted for clinical brain-computer interface applications in paralysed patients. The remainder of this paper describes the experimental methods (Section~\ref{sec:methods}), data records and organisation (Section~\ref{sec:data_records}), technical validation with baseline results (Section~\ref{sec:tech_val}), and usage notes (Section~\ref{sec:usage_notes}), and concludes with data and code availability statements (Sections~\ref{sec:data_availability}--\ref{sec:code_availability}).

\section{Methods}\label{sec:methods} %
\subsection{Participants}

Data were acquired from a single, right-handed male volunteer with normal hearing and vision. A native English speaker, the participant reported no history of neurological, developmental, or language-related disorders. Informed consent was provided by the participant before participating in the experiment, authorising the use of anonymised %
data for research purposes. Ethical approval for the study (R65598/RE001) was provided by the University of Oxford Medical Sciences Interdivisional Research Ethics Committee (IDREC).

\subsection{MEG System \& Recording Setup}

MEG data were acquired at the Oxford Centre for Human Brain Activity (OHBA) using a 306-channel whole-head SQUID MEG system (MEGIN Triux™ Neo, York Instruments Ltd., Heslington, UK). This system, which is housed within a magnetically shielded room to minimise environmental noise, comprises 102 magnetometers and 204 planar gradiometers arranged in 102 triplet sensor units. Data were recorded continuously at a sampling frequency of 1,000 Hz with an online bandpass filter of 0.01--330 Hz, applied during acquisition to remove slow baseline drifts and to prevent aliasing.

Prior to each recording session, the participant's head shape and fiducial locations were digitised using a Polhemus FASTRAK 3D digitiser (Polhemus, Vermont, USA). As fiducial locations, we used left and right preauricular points and the nasion (anatomical landmarks in front of the ears and at the bridge of the nose). Five head position indicator (HPI) coils were attached to the participant's mastoid bones and forehead and driven at distinct sinusoidal frequencies (295, 307, 314, 321, and 328~Hz) via electromagnetic induction to enable continuous head position monitoring throughout the recording. These frequencies are well above the range of neural oscillations typically analysed in MEG (e.g., 0.1--150~Hz). Electrooculogram (EOG) electrodes were placed in a horizontal pair at the outer canthi (corners of the eyes) and a vertical pair above and below the left eye to track eye-related artefacts. Electrocardiogram (ECG) electrodes were located on the clavicle and hip to monitor cardiac activity. 
An electromyogram (EMG) electrode was placed over the jaw to monitor muscle activity and verify compliance with the covert speech instruction for a sample session (session 2, runs 4--6).

Video recordings were collected during the final recording session (session 4, which included the holdout runs 11--12, described below), further confirming the absence of overt articulation. Although video data are not publicly released to preserve participant anonymity, they were available to reviewers. 
EMG recordings, which can also be used to verify covert speech compliance, are included in the public dataset.

The participant was screened for sources of electromagnetic interference such as metallic objects before entering the magnetically shielded room. To minimise displacement and to reduce variability across sessions, the participant's head was positioned as close as possible to a standard reference location in the dewar (the insulated vessel containing the superconducting sensors). The participant was further instructed to remain as still as possible. To record the participant's responses to catch trials, a MEG-compatible button box was employed (ResponsePixx Dual Handheld, VPixx Technologies Inc., Saint-Bruno, Canada). The button box was held in the non-dominant (left) hand to minimise motor activation in the language-dominant hemisphere.

\subsection{Stimulus Materials}

\begin{figure}[tb]
\centering
\includegraphics[width=0.925\textwidth]{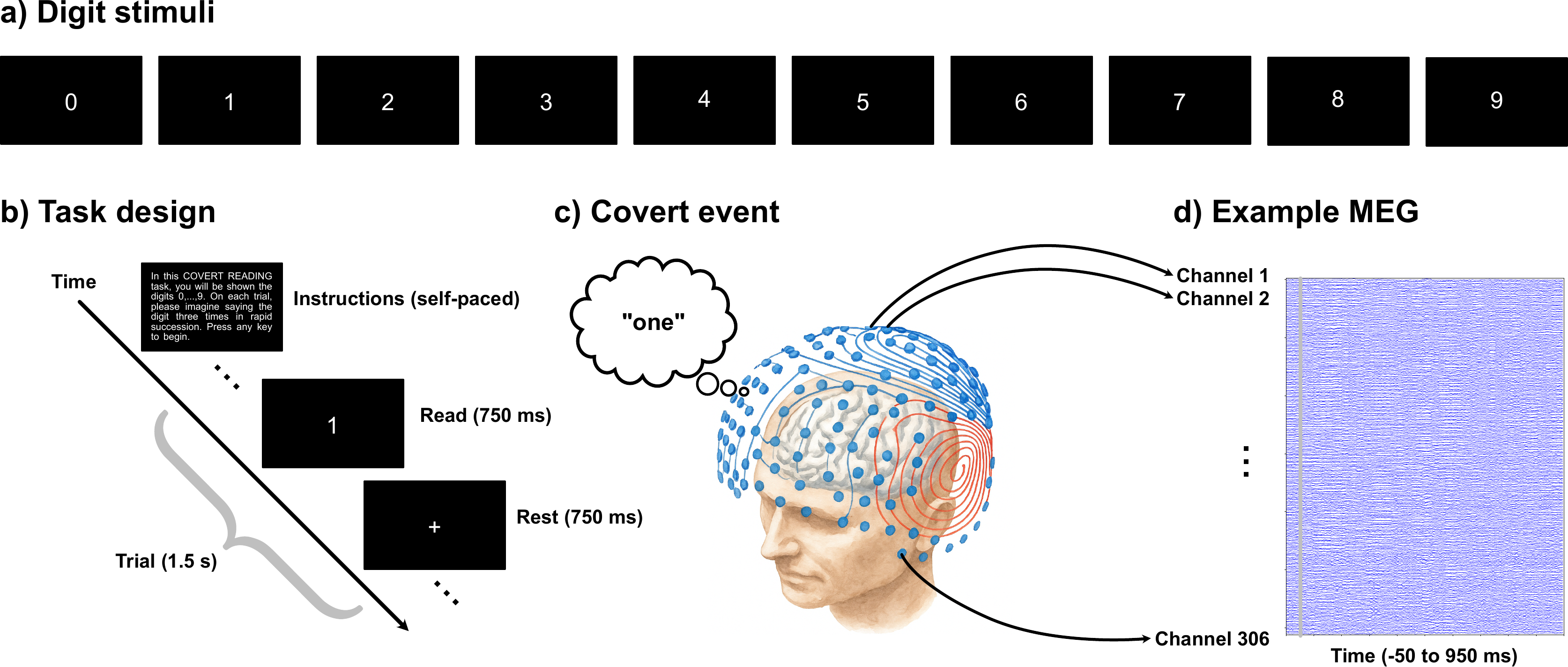}
\caption{MEG digit classification paradigm. (a) Digit stimuli: Visual stimuli consisted of the digits 0--9 presented on a dark background. (b) Task design: Each trial (1.5 s) began with a digit presentation (750 ms) followed by a rest period with fixation cross (750 ms). (c) Covert event: Participants covertly read the presented digit (e.g., ``one'') while MEG signals were recorded from 306 sensors distributed across the scalp. (d) Example MEG: Representative MEG data showing activity across all 306 channels over a 1-second epoch ($-50$ to $950$ ms relative to the stimulus onset, depicted here with a vertical grey line).}
\label{fig:paradigm}
\end{figure}

Visual stimuli consisted of the digit characters 0--9 rendered as text (Figure \ref{fig:paradigm}a). Stimuli were projected onto a translucent whiteboard using a DLP LED projector (ProPixx, VPixx Technologies Inc., Saint-Bruno, Canada), which was positioned in full view of the participant with a 1680 $\times$ 1050 pixel display. Stimulus delivery was managed via the PsychoPy toolbox\cite{peirce2007psychopy} (version 2022.1.4). Digits were displayed as white text in Arial font with a height of 0.2 normalised units against a black background. To synchronise the stimulus computer and MEG system, we used a parallel port to deliver triggers; each trigger marked a stimulus onset with millisecond temporal precision. 

\subsection{Experimental Design \& Procedure}
The experiment used a single-factor design with digit identity (0--9) as the independent variable. Each trial consisted of a digit presented for 750 ms, followed by a fixation cross for 750 ms (inter-stimulus interval), yielding a total trial duration of 1,500 ms (Figure \ref{fig:paradigm}b). The participant was instructed to imagine repeating the digit name three times in rapid succession during digit presentation without making any overt response (Figure \ref{fig:paradigm}c). Three repetitions were chosen as the maximum number that could be comfortably articulated within the 750 ms presentation window, as verified through overt practice trials prior to recording, while providing sustained neural activity throughout the stimulus period. The fixation cross served to maintain fixation and minimise eye movements between trials.

Each run contained 1,000 digit trials evenly distributed across the 10 digit classes (100 trials per digit), presented in pseudorandom order. To monitor alertness, one catch trial was randomly inserted within each sequence of 100 digit trials (10 per run). On catch trials, the word ``button'' replaced the digit stimulus, and the participant pressed a button as quickly as possible during the 750 ms response window. Visual feedback (green text) confirmed successful responses. All catch trials were successfully detected, confirming sustained attention. A self-timed break was provided halfway through each run. Catch trials were excluded prior to epoching, yielding 1,000 digit trials per run.

\begin{figure}[tb]
\centering

\includegraphics[width=1\textwidth]{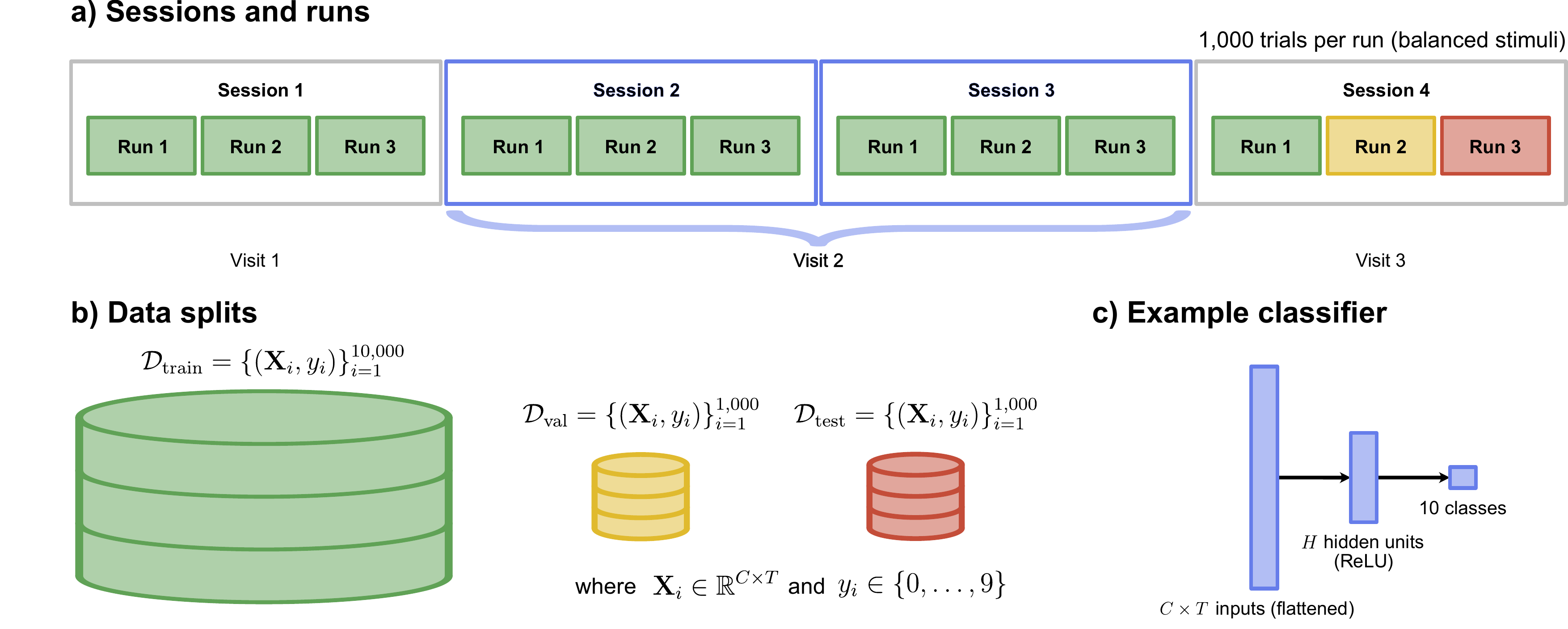}

\caption{Data overview and example usage. (a) Sessions and runs: Data were collected across three visits, comprising four sessions with twelve runs total (1,000 trials per run with balanced digit stimuli); visit 2 encompassed Sessions 2 and 3. (b) Data splits: The dataset was partitioned into training (10,000 examples), validation (1,000 examples), and test (1,000 examples) sets. Formally, each example is a pair $(\mathbf{X}_i, y_i)$, where $\mathbf{X}_i \in \mathbb{R}^{C \times T}$ is a channel-by-time matrix of MEG data (cf.~Figure \ref{fig:paradigm}d) and $y_i \in \{ 0, \dots, 9 \}$ is its corresponding digit label. (c) Example classifier: To illustrate, we use an intentionally simple architecture comprising $306 \times 250$ flattened inputs (76,500 features), $H$ hidden units (1,024 in the best models) with ReLU activation, and 10-class softmax output, as described in the Technical Validation (Section \ref{sec:tech_val}). Formally, the model can be written as $\mathbf{\hat{y}}_i = \mathrm{softmax} (\mathbf{b}_2 + \mathbf{W}_2 \, \mathrm{ReLU} (\mathbf{b}_1 + \mathbf{W}_1 \, \mathrm{vec}(\mathbf{X}_i)))$, where the output $\mathbf{\hat{y}}_i$ is a vector of probabilities over 10 digit classes and the final predicted digit is $\hat{y}_i = \mathrm{arg\,max} \, \hat{\mathbf{y}}_i$.}
\label{fig:sessions_and_splits}
\end{figure}

The experiment comprised four sessions conducted over three non-consecutive visits (over a two-week period): session 1 on visit 1, sessions 2--3 on visit 2, and session 4 on visit 3 (see Figure \ref{fig:sessions_and_splits} for overview). Each session comprised three runs, producing 12 runs and 12,000 digit trials in total. Each session lasted 1--2 hours including setup and breaks.

\subsection{Preprocessing Pipeline}\label{sec:preproc}

Preprocessing was performed using MNE-Python\cite{gramfort2013mne} version 1.4.2 and followed the same minimalist preprocessing approach as the LibriBrain and LibriBrain100 datasets\cite{ozdogan2025libribrain,mantegna2026libribrain100}, to preserve signal characteristics for machine learning applications. The preprocessing pipeline consisted of the following steps:

\textbf{Head position and bad channel identification.} Head position information was extracted from HPI coil measurements using \texttt{compute\_chpi\_amplitudes}, \texttt{compute\_chpi\_locs}, and \texttt{compute\_head\_pos}. Bad channels were identified automatically using 
\texttt{find\_bad\_channels\_maxwell} with a high-pass filter at 40 Hz to remove HPI coil frequencies and line noise, along with calibration and cross-talk files (\texttt{sss\_cal.dat} and \texttt{ct\_sparse.fif}) provided in \texttt{derivatives/neo/}. Identified bad channels were interpolated using signals from neighbouring sensors.

\textbf{Signal space separation.} Signal space separation (SSS)\cite{taulu2005,taulu2006} was applied using \texttt{maxwell\_filter} to suppress external magnetic interference and correct for head movement. A temporally non-extended SSS algorithm was used with origin \texttt{`auto'}, coordinate frame \texttt{`head'}, destination \texttt{(0, 0, 0.04)} to transform data to a standard head position (4 cm above origin), and the calibration and cross-talk files.

\textbf{Filtering.} Notch filters at 50~Hz and 100~Hz removed mains power line interference and its first harmonic. 
A zero-phase Butterworth band-pass filter (0.1--125~Hz) removed slow drifts and prevented aliasing during downsampling.

\textbf{Downsampling and epoching.} Continuous data were downsampled from 1,000 Hz to 250 Hz to reduce computational requirements while preserving signals up to 125 Hz (Nyquist criterion). 
Data were then segmented into epochs spanning $-50$ to 950~ms relative to stimulus onset. The 50~ms pre-stimulus period provides context, while the 950~ms post-stimulus window captured the entire digit presentation period (750~ms), allowing for neural responses extending 200~ms beyond the stimulus offset associated with covert speech imagery.

\textbf{Quality control.} Power spectral density (PSD) was visually inspected after notch filtering, bandpass filtering, and downsampling to verify preprocessing effectiveness. Event timing was verified using the trigger channel (STI101) to ensure accurate stimulus alignment (see \verb|derivatives/preproc/report/|).

\subsection{Data Splits}\label{sec:splits}
The dataset was divided into standardised training, validation, and test splits to enable reproducible benchmarking. Runs 1--10 comprise the training set (10,000 trials), run 11 the validation set (1,000 trials), and run 12 the test set (1,000 trials). 
Rather than randomly assigning trials to data splits, which risks information leakage between train and holdout data from slow (and thus correlated) physiological changes across adjacent trials\cite{harris2020nonsense}, entire experimental runs were assigned to each split. This ensures validation and test sets represent genuinely independent recordings. 
Assigning chronologically later runs to validation and test sets (rather than random selection) %
further reflects realistic brain-computer interface deployment scenarios. 
The splits are provided as HDF5 files (\texttt{train.h5}, \texttt{val.h5}, and \texttt{test.h5}) in \texttt{derivatives/serialised/}, with format details described below.

\subsection{Serialisation}\label{sec:serialisation}
To facilitate machine learning applications, the preprocessed and epoched data for the train, validation, and test splits were converted from FIF to Hierarchical Data Format version 5 (HDF5). Epochs were stored as \texttt{float32} tensors, halving the storage requirements compared to the original \texttt{float64} data in the FIF files while also preserving sufficient numerical precision---\texttt{float32} being standard in deep learning frameworks and hardware optimisation.

Each HDF5 file contains three primary data arrays (also referred to as HDF5 \textit{datasets}): \texttt{data} (MEG signals with shape $N \times 306 \times 250$, representing trials $\times$ channels $\times$ timepoints), \texttt{labels} (digit class labels 0--9 with shape $N$), and \texttt{times} (sample times in seconds, relative to stimulus onset, with shape 250). In each case, the number of trials $N$ varies by data split (10,000 for training, 1,000 for validation, and 1,000 for test).

For convenience, each file also includes HDF5 datasets for the following: \texttt{sensor\_xyz} (3D sensor positions in meters, with shape $306 \times 3$), \texttt{channel\_names} (MEG channel identifiers, with shape 306), and \texttt{channel\_types} (sensor types distinguishing magnetometers from planar gradiometers, with shape 306). These auxiliary datasets enable straightforward sensor-space visualisation and interpretation. For example, users can directly access the spatial location of channel $i$ via \texttt{sensor\_xyz[i]} and its corresponding name via \texttt{channel\_names[i]}. 
Additional metadata are stored as HDF5 \textit{attributes} (file-level metadata). These attributes include sampling frequency (250 Hz), filter parameters (0.1--125 Hz bandpass, 50 and 100 Hz notch), epoch timing (\texttt{tmin} = $-50$ ms, \texttt{tmax} = 950 ms), the preprocessing pipeline identifier (\verb|processing_steps| = \verb|bads+headpos+sss+notch+bp+ds|), event ID mapping (JSON-encoded dictionary mapping digit names to event codes), MNE-Python version, and a summary of bad channels identified during preprocessing for each run prior to interpolation (\verb|bad_channels_by_run|).

\section{Data Records}\label{sec:data_records}

The MegNIST dataset is organised according to the Brain Imaging Data Structure (BIDS) specification\cite{gorgolewski2016bids} for MEG data\cite{niso2018meg-bids} and was prepared using MNE-BIDS\cite{appelhoff2019mne-bids}. It is publicly available on Hugging Face (\url{https://huggingface.co/datasets/pnpl/MegNIST/}), with the top-level directory structure shown in Figure~\ref{fig:directory-structure}.

\begin{figure}[t]
\centering
\setlength{\fboxsep}{0pt}
\fbox{\colorbox{gray!10}{%
\begin{minipage}{\textwidth}
{\small
\dirtree{%
.1 MegNIST/.
.2 participants.json.
.2 participants.tsv.
.2 dataset\_description.json.
.2 README.
.2 derivatives/.
.2 sub-0/.
.3 ses-1/.
.4 meg/.
.5 sub-0\_ses-1\_task-MegNIST\_run-1\_channels.tsv.
.5 sub-0\_ses-1\_task-MegNIST\_run-1\_events.tsv.
.5 sub-0\_ses-1\_task-MegNIST\_run-1\_meg.fif.
.5 sub-0\_ses-1\_task-MegNIST\_run-1\_meg.json.
.5 sub-0\_ses-1\_task-MegNIST\_run-2\_channels.tsv.
.5 sub-0\_ses-1\_task-MegNIST\_run-2\_events.tsv.
.5 sub-0\_ses-1\_task-MegNIST\_run-2\_meg.fif.
.5 sub-0\_ses-1\_task-MegNIST\_run-2\_meg.json.
.5 sub-0\_ses-1\_task-MegNIST\_run-3\_channels.tsv.
.5 sub-0\_ses-1\_task-MegNIST\_run-3\_events.tsv.
.5 sub-0\_ses-1\_task-MegNIST\_run-3\_split-01\_meg.fif.
.5 sub-0\_ses-1\_task-MegNIST\_run-3\_split-02\_meg.fif.
.5 sub-0\_ses-1\_task-MegNIST\_run-3\_meg.json.
.5 sub-0\_ses-1\_coordsystem.json.
.4 sub-0\_ses-1\_scans.tsv.
.3 ses-2/.
.4 meg/.
.5 \ldots.
.4 \ldots.
.3 ses-3/.
.4 meg/.
.5 \ldots.
.4 \ldots.
.3 ses-4/.
.4 meg/.
.5 \ldots.
.4 \ldots.
}
}
\end{minipage}%
}}
\caption{Directory structure of the MegNIST dataset following BIDS conventions.}
\label{fig:directory-structure}
\end{figure}

The MegNIST dataset contains raw MEG recordings from a single participant (\texttt{sub-0}) across four sessions (\texttt{ses-1}, \texttt{ses-2}, \texttt{ses-3}, \texttt{ses-4}). Each session includes three experimental runs (\texttt{run-1}, \texttt{run-2}, \texttt{run-3}), with each run containing 1,000 trials evenly distributed across the 10 digit classes from the MNIST dataset\cite{lecun1998mnist} (100 trials per class). The raw MEG data are stored in FIF format (\texttt{.fif} files), the native format of Elekta-Neuromag systems, with each \texttt{.fif} file accompanied by a BIDS-compliant JSON sidecar file (\texttt{.json}) and a tab-separated value file (\texttt{.tsv}). The \texttt{.tsv} files list channel specifications, while the \texttt{.json} files contain acquisition parameters and metadata.

At the root level, the dataset includes standard BIDS metadata files. The \texttt{participants.tsv} and \texttt{participants.json} files contain the BIDS-compliant participant metadata structure, with demographic details anonymised for a single participant. The \texttt{dataset{\textunderscore}description.json} file contains dataset-level metadata including the dataset name, dataset type, and BIDS version (1.7.0). The \texttt{README} file provides references to the MEG-BIDS specification\cite{niso2018meg-bids} and MNE-BIDS software\cite{appelhoff2019mne-bids} used to create the dataset.

Within each session directory (\texttt{sub-0/ses-*/meg/}), the raw MEG recordings are organised by run. Each run is represented by a primary \texttt{.fif} file containing the continuous MEG data sampled at 1,000 Hz. The recordings include 306 MEG channels (102 magnetometers and 204 planar gradiometers), 2 electrooculogram (EOG) channels, 1 electrocardiogram (ECG) channel, and additional reference and trigger channels. 
A single electromyogram (EMG) channel was recorded over the right masseter muscle to monitor jaw muscle activity and verify the absence of overt articulation for a subset of covert speech trials (session 2, runs 4--6). To enable continuous head position tracking via electromagnetic induction, 
HPI coils were driven at five fixed sinusoidal frequencies between 295 and 328~Hz---these frequencies lie well above the 250~Hz Nyquist limit implied by our 125~Hz low-pass filter (see Section~\ref{sec:preproc}). Each run also contains a \texttt{.json} sidecar with acquisition parameters (including sampling frequency, hardware filters, power line frequency, and task parameters) and a \texttt{channels.tsv} file listing all sensors with their name, type, units, hardware filter settings, and status. Some runs are split across multiple files (e.g., \texttt{split-01}, \texttt{split-02}) due to file size constraints of the acquisition system. Additionally, each session includes a \texttt{coordsystem.json} file that specifies the coordinate system used for MEG sensor positions and digitised head landmarks, enabling spatial alignment across sessions. %

\begin{figure}[t]
\centering
\setlength{\fboxsep}{0pt}
\fbox{\colorbox{gray!10}{%
\begin{minipage}{\textwidth}
{\small
\dirtree{%
.1 derivatives/.
.2 neo/.
.3 sss\_cal.dat.
.3 ct\_sparse.fif.
.3 sensor\_xyz.json.
.2 preproc/.
.3 headpos/.
.4 sub-0\_ses-1\_task-MegNIST\_run-1.csv.
.4 \ldots.
.4 sub-0\_ses-4\_task-MegNIST\_run-3.csv.
.3 sub-0/.
.4 ses-1/.
.5 meg/.
.6 sub-0\_ses-1\_task-MegNIST\_run-1\_bad\_channels.json.
.6 sub-0\_ses-1\_task-MegNIST\_run-1\_proc-bads+headpos+sss\_meg.fif.
.6 sub-0\_ses-1\_task-MegNIST\_run-1\_proc-bads+headpos+sss+notch+bp+ds\_meg.fif.
.6 sub-0\_ses-1\_task-MegNIST\_run-1\_proc-bads+headpos+sss+notch+bp+ds\_epo.fif.
.6 \ldots.
.6 sub-0\_ses-1\_task-MegNIST\_run-3\_proc-bads+headpos+sss+notch+bp+ds\_epo.fif.
.4 ses-2/.
.5 meg/.
.6 \ldots.
.4 ses-3/.
.5 meg/.
.6 \ldots.
.4 ses-4/.
.5 meg/.
.6 \ldots.
.3 report/.
.4 sub-0\_ses-1\_task-MegNIST\_run-1\_\ldots\_meg.html.
.4 \ldots.
.4 sub-0\_ses-4\_task-MegNIST\_run-3\_\ldots\_meg.html.
.2 serialised/.
.3 train.h5.
.3 val.h5.
.3 test.h5.
}
}
\end{minipage}%
}}
\caption{Structure of the derivatives folder containing preprocessed data and serialised splits.}
\label{fig:derivatives-structure}
\end{figure}

The \texttt{derivatives/} folder contains preprocessed data and analysis outputs, with its structure shown in Figure~\ref{fig:derivatives-structure}. While BIDS permits derivatives folders, it does not strictly define their contents, allowing flexibility for dataset-specific processing pipelines. The preprocessing steps and derivatives are described in more detail in the Methods (Section \ref{sec:preproc}). 
Briefly, the \texttt{neo/} subdirectory contains calibration and cross-talk correction files (\texttt{sss\_cal.dat} and \texttt{ct\_sparse.fif}) required for signal space separation (SSS)\cite{taulu2005,taulu2006}, a noise reduction technique implemented here using MNE-Python\cite{gramfort2013mne}. 
The \texttt{neo/} subdirectory also contains a file, \texttt{sensor\_xyz.json}, which stores the 3D MEG sensor positions that populate the \texttt{sensor\_xyz} arrays in the serialised HDF5 files. %
The \texttt{preproc/} subdirectory contains: head position estimates (\texttt{headpos/}), quality control reports (\texttt{report/}), and preprocessed MEG data. The preprocessed MEG directory mirrors the raw data structure, where, for each session (e.g., \texttt{sub-0/ses-1/meg/}), there are preprocessed files for runs 1--3. For each run, we have cached files with progressive amounts of preprocessing (for details, see Section \ref{sec:preproc}). Preprocessed MEG files with the suffix \texttt{proc-bads+headpos+sss\_meg.fif} had bad channels identified (\texttt{bads}), headposition corrected (\texttt{headpos}), and noise removed from outside of the head (\texttt{sss}). Additional notch filtering (\texttt{notch}), bandpass filtering (\texttt{bp}), and downsampling (\texttt{ds}) have been applied to files suffixed with \texttt{proc-bads+headpos+sss+notch+bp+ds\_meg.fif}. Files that end with \texttt{epo.fif}, rather than \texttt{meg.fif}, have been epoched. 

The \texttt{serialised/} subdirectory contains predefined training (\texttt{train.h5}), validation (\texttt{val.h5}), and test (\texttt{test.h5}) splits in HDF5 format. These \texttt{.h5} files store the data as preprocessed \texttt{float32} tensors, reducing file size by more than half and removing the need for domain-specific expertise in neuroimaging file structures, preprocessing, and software libraries. This allows researchers from machine learning backgrounds to work directly with the data using familiar machine learning tools (e.g., PyTorch\cite{paszke2019pytorch}). The standardised data splits support reproducible benchmarking and fair comparison of different methods, following foundational machine learning datasets like MNIST\cite{lecun1998mnist} and CIFAR-10\cite{krizhevsky2009cifar10}.

\section{Technical Validation}\label{sec:tech_val}

To validate that the MegNIST dataset contains decodable information about imagined digits, we performed baseline classification experiments using a simple neural network model. This validation serves two purposes: first, to demonstrate that neural signals recorded during inner speech contain digit-related information significantly above chance levels; second, to establish a reproducible baseline for future method comparisons.

\subsection{Decoding Approach}\label{sec:decoding}

We adopted a full-epoch decoding approach\cite{csaky2023interpretable}, where the entire temporal window of each trial is used for classification. Each MEG recording (306 channels $\times$ 250 timepoints) was flattened into a single feature vector of 76,500 elements. This straightforward approach avoids making assumptions about the temporal dynamics of inner speech and provides a simple baseline that future methods can build upon.
To demonstrate digit classification from brain recordings, we use a simple multilayer perceptron (MLP). This model has a single architectural hyperparameter $H$, which represents the number of units in the hidden layer. In total, we evaluated seven hidden layer sizes increasing by powers of 2 from 32 to 2,048 units. Formally, for a given input trial $\mathbf{X}_i \in \mathbb{R}^{306 \times 250}$ from the dataset, the model computes:
\begin{align*}
\mathbf{\hat{y}}_i = \mathrm{softmax}(\mathbf{b}_2 + \mathbf{W}_2 \, \mathrm{ReLU}(\mathbf{b}_1 + \mathbf{W}_1 \, \mathrm{vec}(\mathbf{X}_i))).
\end{align*}
Here $\mathrm{vec}(\mathbf{X}_i) \in \mathbb{R}^{76{,}500}$ represents the flattened input. 
The model parameters have the following dimensions: 
$\mathbf{W}_1 \in \mathbb{R}^{H \times 76{,}500}$, $\mathbf{b}_1 \in \mathbb{R}^H$, $\mathbf{W}_2 \in \mathbb{R}^{10 \times H}$, $\mathbf{b}_2 \in \mathbb{R}^{10}$. 
The vector $\mathbf{\hat{y}}_i \in \mathbb{R}^{10}$ is the predicted probability distribution over digit classes, where the predicted digit itself is given by $\hat{y}_i = \mathrm{arg\,max} \, \hat{\mathbf{y}}_i$. 
A schematic of the model is presented in Figure \ref{fig:sessions_and_splits}c.

Models were trained using the AdamW optimiser\cite{loshchilov2019AdamW} with cross-entropy loss and early stopping  to prevent overfitting based on validation accuracy (we used a patience of 20 epochs). The implementation used PyTorch\cite{paszke2019pytorch} with common hyperparameters for optimisation: learning rate 0.001, batch size 32, and a maximum of 200 epochs.
To enable robust model selection and paired statistical comparisons, we trained each configuration with ten random seeds (repeated across all architectures). This approach accounts for training variability while increasing statistical power by enabling paired analyses that control for seed-specific variance. All data were standardised (zero mean, unit variance) using parameters computed on the training set and applied to validation and test sets to prevent information leakage. 
Within-run standardisation, in which each run was standardised using its own statistics, was found to yield slightly higher accuracy in exploratory analyses.

\subsection{Model Selection and Evaluation}\label{sec:model_selection}

\begin{figure}[htbp]
\centering

\includegraphics[width=1\textwidth]{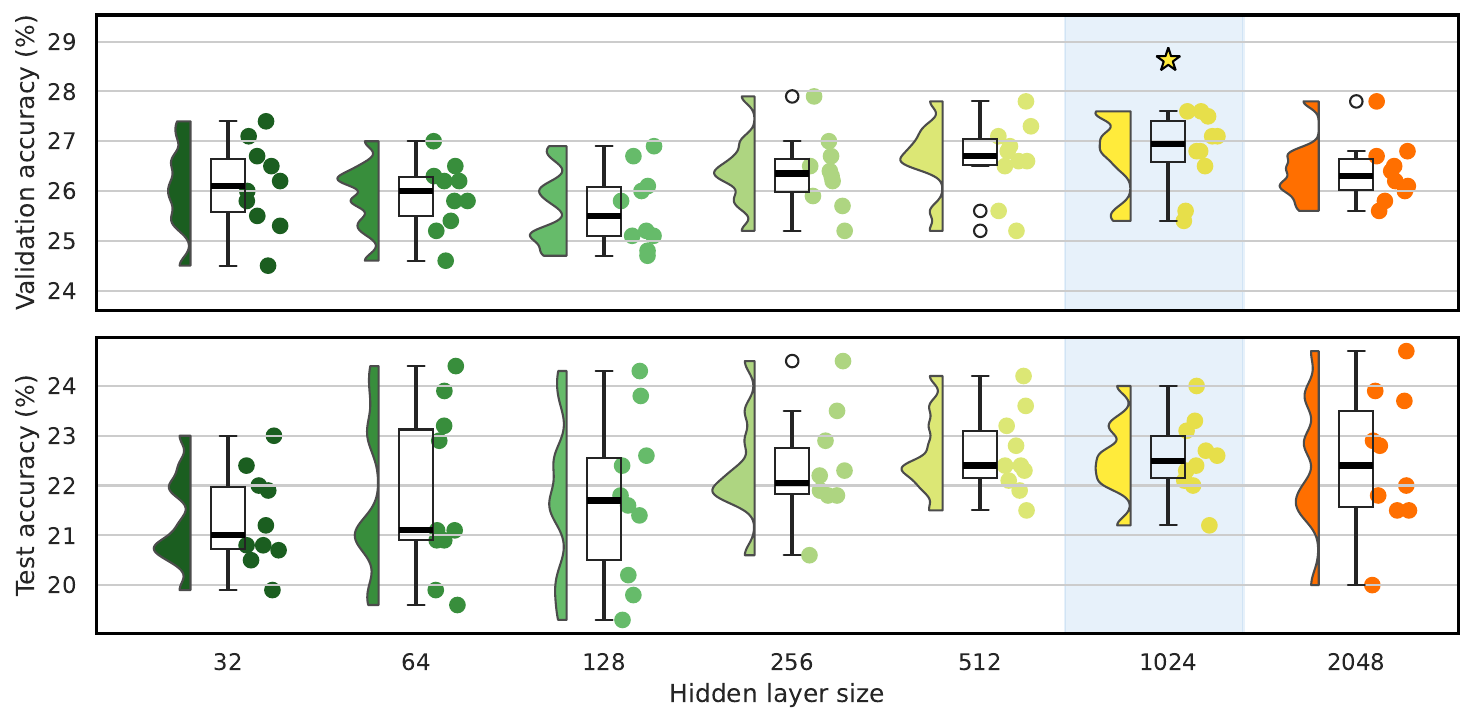}

\caption{Model selection and generalisation. Validation (top) and 
test (bottom) accuracy across hidden layer sizes (10 seeds each). 
Each architecture is shown with individual data points, 
box plots (median as bold line, interquartile range as box, whiskers 
at 1.5 $\times$ IQR), and kernel density estimates (half-violins). Colours progress from green (smaller models) through yellow (optimal) 
to orange (larger models). The 1024-unit architecture (yellow star, %
blue shading) 
achieved the highest validation performance (26.8 ± 0.8\%) %
and was selected. 
Test accuracy confirms reliable generalisation. Both panels use matching 
y-axis spans for direct comparison.}
\label{fig:val_test_raincloud}
\end{figure}

We selected the optimal architecture based on mean validation accuracy across the 10 random seeds. Figure~\ref{fig:val_test_raincloud} shows validation and test accuracy distributions for each hidden layer size. The 1024-unit architecture achieved the highest mean validation accuracy (26.8 ± 0.8\%) and was selected for downstream analyses. Test accuracy for this architecture (22.6 ± 0.8\% across 10 seeds) closely tracked validation performance, substantially exceeding the 10\% chance level expected for balanced 10-class classification.

We needed a single representative model for a post-hoc interpretability analysis (Section~\ref{sec:posthoc-interpretability}). To identify it, we selected the random seed whose validation accuracy was closest to the median across the 10 runs. Four seeds were equally close to the validation median. We therefore selected the one whose test accuracy was closest to the median (i.e., seed 47), rather than favouring the seed with the highest test accuracy (as that would have favoured an unusually lucky run). We emphasise that this tiebreaker did not affect architecture selection or the reported mean test performance. For comparison, mean validation and test accuracies across the 10 seeds were 26.8 ± 0.8\% and 22.6 ± 0.8\%, respectively.

\subsection{Statistical Significance Testing}

We evaluated whether the observed decoding performance exceeded chance by performing permutation hypothesis testing\cite{good2000, ojala2010}. This non-parametric approach constructs an empirical null distribution by repeatedly evaluating the trained classifier on randomly permuted labels, thereby estimating the probability of observing the observed test accuracy under the null hypothesis that digit labels are unrelated to neural signals.

To construct the null distribution, we randomly permuted the 1,000 test labels 100,000 times, recording the classification accuracy after each permutation while holding fixed the predictions for the representative model (using seed 47, introduced in Section~\ref{sec:model_selection}). The resulting null distribution represents accuracies obtained under randomly reassigned labels. We estimated the one-tailed empirical $p$-value using the standard $+1$ correction, $p=(b+1)/(100000+1)$, where $b$ is the number of permuted accuracies greater than or equal to the observed accuracy. This indicates how rarely performance at least as high as the observed test accuracy occurs under the null hypothesis.

\begin{figure}[htbp]
\centering

\includegraphics[width=1\textwidth]{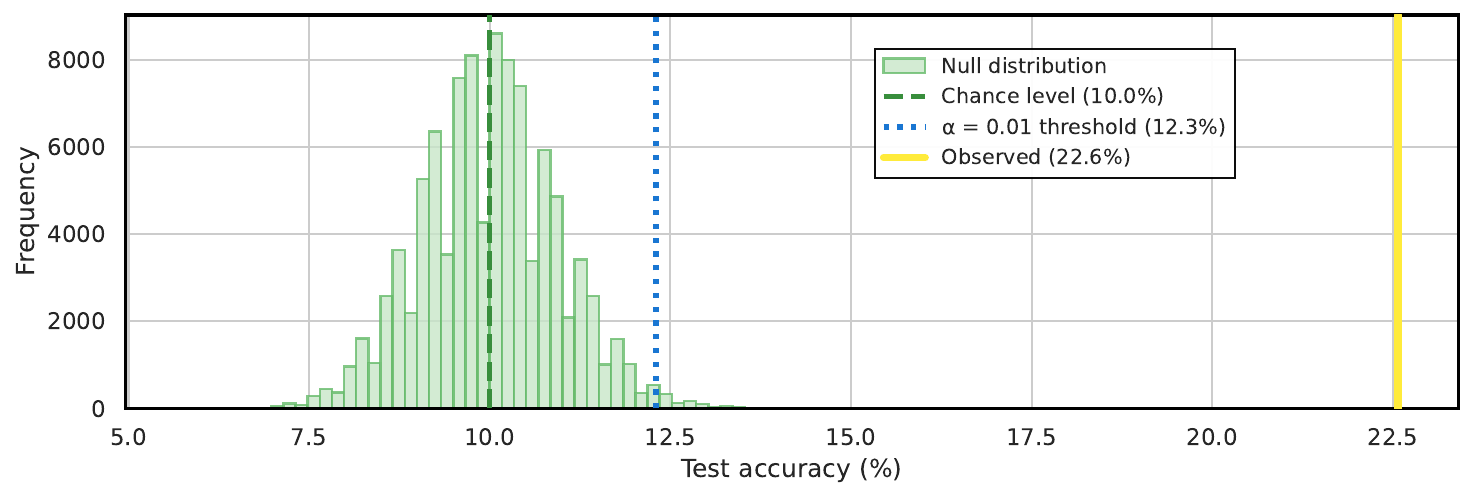}

\caption{Decoding performance. Using 100,000 test-label permutations, the observed test accuracy of 22.6\% (yellow line) was significantly higher than expected under the null distribution (green histogram), exceeding it by a wide margin: the dashed green line shows chance (10\%) for balanced 10-class classification, and the dotted blue line marks the critical value of $12.3\%$, which is the smallest attainable accuracy with $p \leq 0.01$.}
\label{fig:permutation_vs_chance}
\end{figure}

Figure~\ref{fig:permutation_vs_chance} shows the null distribution as a histogram with the true test accuracy marked, demonstrating that observed performance is highly statistically significant ($p \ll 0.01$). 
Here we adopt a conventional significance threshold of $\alpha = 0.01$. The observed test accuracy of 22.6\% yielded an empirical $p$-value of $1/100001 \approx 1.0 \times 10^{-5}$. The null distribution had a median of 10.0\% (expected for balanced 10-class classification) and a 99th percentile of 12.2\%. Because accuracy is discrete in increments of 0.1 percentage points, the smallest attainable accuracy with a one-tailed empirical $p \leq 0.01$ was 12.3\%.

These results confirm that the MegNIST dataset contains robust, decodable information about digit class that can be reliably extracted using simple neural network methods. The significant above-chance performance validates the dataset's utility for machine learning research and establishes a baseline for future method development.

\subsection{Post-hoc Interpretability Analysis}\label{sec:posthoc-interpretability}

To understand which temporal and spatial features contributed most to decoding performance, we performed permutation feature importance analysis on the trained MLP. 
This analysis identifies which input features are most informative for classification by measuring the decrease in accuracy when sensors are randomly shuffled. 
Concretely, we slid a $\pm 50$~ms window across all 250 timepoints of the trial epoch, randomly permuting sensors within each sliding window and computing the mean decrease in accuracy across 10 repeats. %

\begin{figure}[htbp]
\centering

\includegraphics[width=1\textwidth]{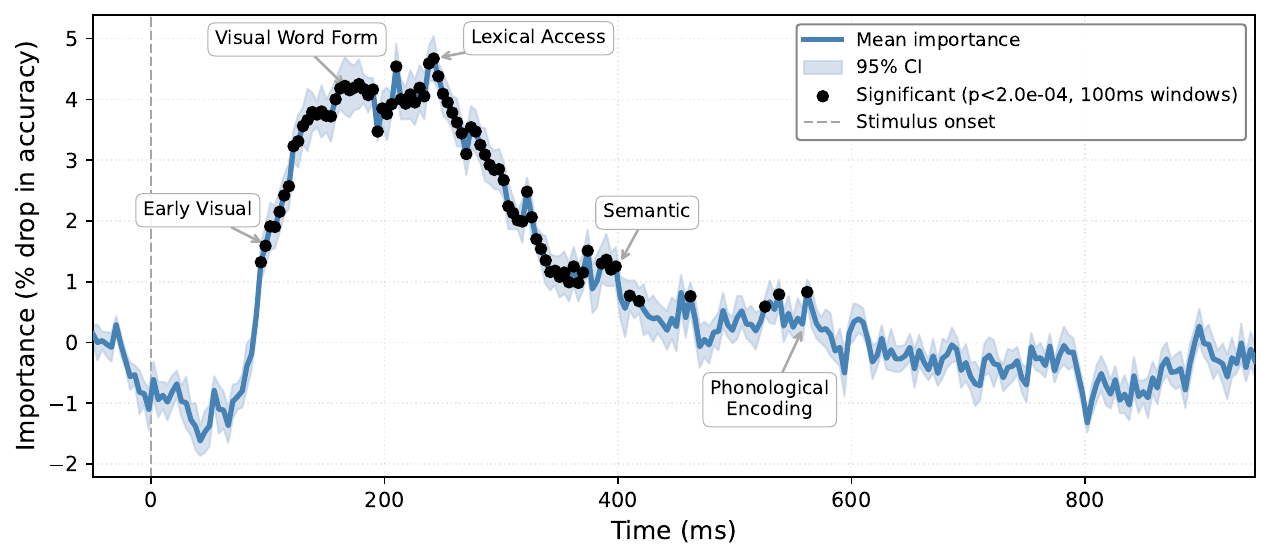}

\caption{Permutation feature importance (PFI) in time during imagined-speech decoding. The y-axis shows the \% drop in accuracy when spatial features are shuffled within a $\pm50$~ms window around each time point. More important times show higher values, with shaded regions providing 95\% confidence intervals and black dots indicating significant importance (one-sided one-sample $t$-tests across 10 repeats, Bonferroni-corrected; $p < 2 \times 10^{-4}$). Importance rises sharply following stimulus onset, peaking at 242 ms (4.8\% drop) which is a response time associated with linguistic processing\cite{pylkkanen2003,salmelin2007}. Earlier importance (100--170 ms) corresponds to visual and orthographic processing\cite{tarkiainen1999,cohen2002}, while later sustained importance (300–600 ms) reflects semantic and phonological encoding\cite{kutas2011,indefrey2011,tian2012mental}. Negative values indicate random variability where permutation slightly improved accuracy relative to the baseline.}
\label{fig:temporal_pfi}
\end{figure}

As shown in Figure~\ref{fig:temporal_pfi}, decoding importance peaked at approximately 242 ms post-stimulus onset, corresponding to the M250 component typically associated with lexical access\cite{pylkkanen2003, salmelin2007}. Earlier peaks around 100 and 170 ms correspond to the M100 and M170 components linked to early visual and visual-word-form processing\cite{tarkiainen1999, cohen2002}. Later, more sustained importance between 350--600 ms is consistent with semantic and phonological encoding stages\cite{kutas2011,indefrey2011,tian2012mental}.

\begin{figure}[htbp]
\centering

\includegraphics[width=1\textwidth]{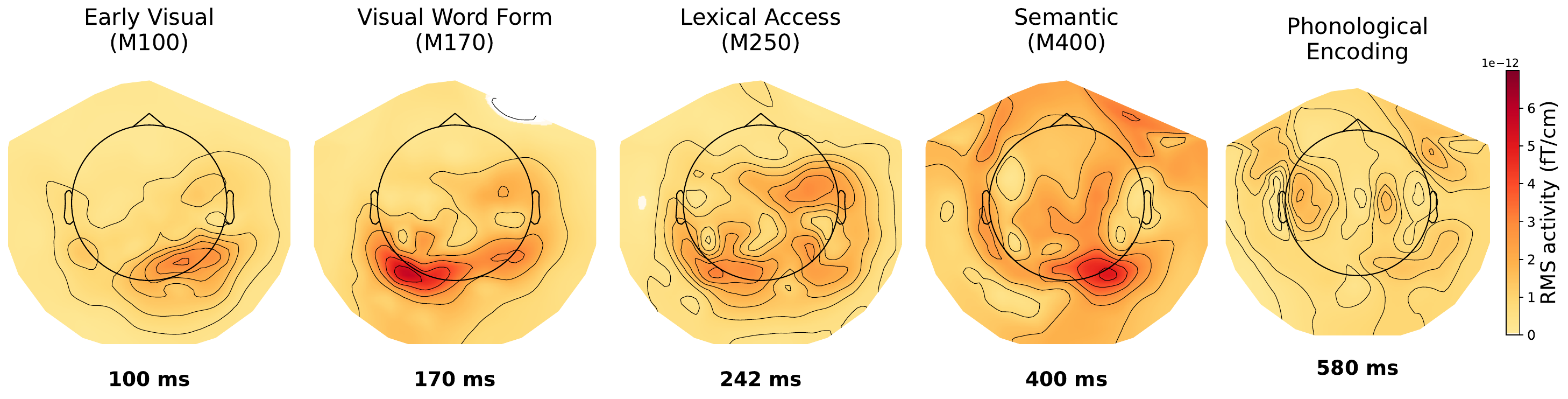}

\caption{Spatial–temporal progression of neural activity during digit recognition with imagined speech. Baseline-corrected planar-gradiometer RMS topographies at five key latencies illustrate the cascade from early visual (M100) and visual-word-form (M170) processing through lexical-semantic (M250–M400) to phonological encoding ($\sim$580 ms). All maps share a common scale for direct comparison.}
\label{fig:topomaps}
\end{figure}

To visualise the spatial organisation of these processes, we computed topographic maps of the baseline-corrected planar-gradiometer RMS activity averaged across all trials. %
The results in Figure~\ref{fig:topomaps} illustrate a clear posterior-to-anterior progression of activity: occipital visual responses ($\sim$100 ms) are followed by occipito-temporal engagement of the visual-word-form system ($\sim$170 ms), then by distributed temporal–parietal–frontal activation near the time of peak decoding importance ($\sim$242 ms). Activity at 400 ms shows widespread temporal and frontal involvement consistent with semantic processing, while later ($\sim$580 ms) frontal-central patterns align with phonological encoding and articulatory imagery.

These post-hoc analyses provide insight into the spatiotemporal dynamics of inner speech processing captured in the MegNIST dataset, demonstrating that the neural signals contain structured, interpretable patterns that extend beyond simple statistical separability.

\section{Usage Notes}\label{sec:usage_notes}

A minimal example demonstrating the preprocessing pipeline for the raw data is provided below in Python.

\begin{lstlisting}[language=Python, caption={Import MNE-Python and load raw MEG and events}, label={lst:preproc}]
import mne

raw = mne.io.read_raw_fif('sub-0_ses-1_task-MegNIST_run-1_meg.fif', preload=True)
events = mne.find_events(raw, stim_channel='STI101', min_duration=0.005, mask=255)

event_id = {
    'digit/zero': 10,
    'digit/one': 11,
    'digit/two': 12,
    'digit/three': 13,
    'digit/four': 14,
    'digit/five': 15,
    'digit/six': 16,
    'digit/seven': 17,
    'digit/eight': 18,
    'digit/nine': 19,
}
\end{lstlisting}

\begin{lstlisting}[language=Python, , firstnumber=last, caption={MNE-Python preprocessing pipeline}, label={lst:preproc}]
# Get head positions
chpi_amplitudes = mne.chpi.compute_chpi_amplitudes(raw)
chpi_locs = mne.chpi.compute_chpi_locs(raw.info, chpi_amplitudes)
head_pos = mne.chpi.compute_head_pos(raw.info, chpi_locs)

# Find bad channels
auto_noisy_chs, auto_flat_chs, auto_scores = mne.preprocessing.find_bad_channels_maxwell(raw, calibration='sss_cal.dat', cross_talk='ct_sparse.fif', h_freq=40, picks='meg', return_scores=True)
raw.info['bads'] = auto_noisy_chs + auto_flat_chs

# Maxfilter
raw = mne.preprocessing.maxwell_filter(raw, origin='auto', coord_frame='head', destination=(0, 0, 0.04), calibration='sss_cal.dat', cross_talk='ct_sparse.fif', head_pos=head_pos)

# Notch filter
raw.notch_filter(freqs=[50, 100], picks='meg')

# Bandpass filter
raw.filter(l_freq=0.1, h_freq=125)

# Downsample
raw, events = raw.resample(250, events=events)

# Epoch
epochs = mne.Epochs(raw, events, event_id, -0.050, 0.950, baseline=None, preload=True, reject_by_annotation=False).pick('meg') 
epochs.crop(tmax=epochs.times[-2])  # 250 samples over interval of [-50, 950) ms
\end{lstlisting}

\begin{lstlisting}[language=Python, , firstnumber=last, caption={Save preprocessed epochs   }, label={lst:preproc}]
# Save epochs file
epochs.save('sub-0_ses-1_task-MegNIST_run-1_proc-bads+headpos+sss+notch+bp+ds_epo.fif', overwrite=True)

\end{lstlisting}

\begin{lstlisting}[language=Python, , firstnumber=last, caption={Get data and labels}, label={lst:preproc}]
X = epochs.get_data()           # (n_epochs, n_channels, n_times)
y = epochs.events[:, 2] - 10    # integers 0..9
\end{lstlisting}

\section{Data Availability}\label{sec:data_availability}

All data described in this manuscript are publicly available on Hugging Face in two formats:
\begin{itemize}
    \item a full BIDS-compliant version (\url{https://huggingface.co/datasets/pnpl/MegNIST/tree/main}), and
    \item a serialised HDF5 version optimised for programmatic access (\url{https://huggingface.co/datasets/pnpl/MegNIST/tree/main/derivatives/serialised}).
\end{itemize}

\section{Code Availability}\label{sec:code_availability}

To enable easy replication of the data preprocessing, splitting, and serialisation described above (see Sections~\ref{sec:preproc}, \ref{sec:splits}, and \ref{sec:serialisation}),
all code used to process the dataset from raw MEG recordings and reproduce the Technical Validation analyses and figures is available at \url{https://github.com/neural-processing-lab/megnist-paper}. 
Random seeds used in stochastic analyses are specified explicitly in the released code to facilitate reproducibility. The code uses standard open-source scientific Python packages, including \texttt{MNE-Python}~\cite{gramfort2013mne} for MEG-specific preprocessing and \texttt{PyTorch}\cite{paszke2019pytorch} for the MLP model. 

MegNIST is optionally accessible through the \texttt{pnpl} library\cite{landau2025competition} for streamlined data access and preprocessing. The \texttt{pnpl} library can be installed directly from the command line using \texttt{pip} (\texttt{pip install pnpl}). In Python, the \texttt{pnpl} library will automatically download the requested MegNIST data split (\texttt{train.h5}, \texttt{val.h5}, \texttt{test.h5}) if it is not already present on the specified path. 

\begin{lstlisting}[language=Python, caption={MegNIST dataloading is easy using the \texttt{pnpl} library in Python}, label={lst:pnpl}]
from pnpl.datasets import MegNIST
from torch.utils.data import DataLoader

train = MegNIST("./data/MegNIST", partition="train") 
val   = MegNIST("./data/MegNIST", partition="val")
test  = MegNIST("./data/MegNIST", partition="test")

loader = DataLoader(train, batch_size=32, shuffle=True)
x, y = next(iter(loader))  # x: (32, 306, 250), y: (32,)
\end{lstlisting}

\bibliography{references}
\bibliographystyle{naturemag} %

\section{Acknowledgements}

This work is dedicated to two humble model systems with outsized influence: MNIST and the lesser fruit fly (\textit{Drosophila melanogaster}). The authors acknowledge the NVIDIA Corporation and the University of Oxford Advanced Research Computing (ARC) facility (http://dx.doi.org/10.5281/zenodo.22558) for providing computational resources.

\section{Funding}

Data collection was generously funded by a Major Research Grant from Jesus College, Oxford. PNPL is supported by the MRC (MR/X00757X/1), Royal Society (RG\textbackslash R1\textbackslash 241267), NSF (2314493), NFRF (NFRFT-2022-00241), SSHRC (895-2023-1022), Encode: AI for Science, and ARIA (SCNI-SE01-P004).

\section{Author Contributions}
Experimental design: O.P.J. Data collection and quality control: A.C., S.B., O.P.J. Software engineering: P.S., G.E., O.P.J. Technical validation: F.M., O.P.J. Writing -- original draft: O.P.J. Writing -- review and editing: T.K., P.S., G.E., F.M., A.C., S.B., M.W., O.P.J. Funding: O.P.J.

\section{Competing Interests}
The authors declare no competing interests.

\end{document}